\documentclass[sigconf,nonacm,screen]{acmart}

\usepackage{booktabs}
\usepackage{multirow}
\usepackage{makecell}
\usepackage{adjustbox}

\newcommand{\tablestyle}[2]{\setlength{\tabcolsep}{#1}\renewcommand{\arraystretch}{#2}}

\AtBeginDocument{%
  }

\title{PURE: Refining Feed-forward 3D Reconstruction with Unlabeled Post-training}
\titlenote{Author's version of the paper accepted to the 34th ACM International Conference on Multimedia (MM~'26), November 10--14, 2026, Rio de Janeiro, Brazil. The Version of Record is \url{https://doi.org/10.1145/3767308.3836296}.}

\author{Yaolong Zhu}
\email{yaolongzhu25@stu.pku.edu.cn}
\affiliation{%
  \institution{Guangdong Provincial Key Laboratory of Ultra High Definition Immersive Media Technology, Shenzhen Graduate School, Peking University}
  \city{Shenzhen}
  \country{China}}

\author{Peng Li}
\email{plibp@connect.ust.hk}
\affiliation{%
  \institution{The Hong Kong University of Science and Technology}
  \city{Hong Kong}
  \country{China}}

\author{Jiahao Lu}
\authornote{Project lead.}
\email{jlucn@connect.ust.hk}
\affiliation{%
  \institution{The Hong Kong University of Science and Technology}
  \city{Hong Kong}
  \country{China}}

\author{Chengfeng Zhao}
\email{chengfeng.zhao@connect.ust.hk}
\affiliation{%
  \institution{The Hong Kong University of Science and Technology}
  \city{Hong Kong}
  \country{China}}

\author{Mengfei Li}
\email{mliek@connect.ust.hk}
\affiliation{%
  \institution{The Hong Kong University of Science and Technology}
  \city{Hong Kong}
  \country{China}}

\author{Chao Wang}
\email{winchao1984@gmail.com}
\affiliation{%
  \institution{Pengcheng Laboratory}
  \city{Shenzhen}
  \country{China}}

\author{Wenyue Chen}
\email{2501212822@stu.pku.edu.cn}
\affiliation{%
  \institution{Guangdong Provincial Key Laboratory of Ultra High Definition Immersive Media Technology, Shenzhen Graduate School, Peking University}
  \city{Shenzhen}
  \country{China}}

\author{Zekai Gu}
\email{zekai.gu@connect.ust.hk}
\affiliation{%
  \institution{The Hong Kong University of Science and Technology}
  \city{Hong Kong}
  \country{China}}

\author{Chengwei Ren}
\email{chernweiren@gmail.com}
\affiliation{%
  \institution{The Hong Kong University of Science and Technology}
  \city{Hong Kong}
  \country{China}}

\author{Ronggang Wang}
\correspondingauthor
\authornote{Co-corresponding authors.}
\email{rgwang@pkusz.edu.cn}
\affiliation{%
  \institution{Guangdong Provincial Key Laboratory of Ultra High Definition Immersive Media Technology, Shenzhen Graduate School, Peking University}
  \city{Shenzhen}
  \country{China}}

\author{Yuan Liu}
\correspondingauthor
\authornotemark[3]
\email{yuanly@ust.hk}
\affiliation{%
  \institution{The Hong Kong University of Science and Technology}
  \city{Hong Kong}
  \country{China}}

\renewcommand{\shortauthors}{Zhu et al.}

\begin{abstract}
Feed-forward 3D reconstruction models have achieved remarkable progress through large-scale training and strong learned priors.
However, they remain fragile when applied to dynamic in-the-wild videos, where non-rigid motion, occlusions, and illumination changes are prevalent.
A key limitation lies in their reliance on expensive depth and camera pose annotations, which hinders scaling to diverse real-world dynamic data and restricts generalization.
To address this issue, we propose PURE, an unlabeled post-training refinement framework that improves depth and camera estimation without requiring ground-truth supervision.
Instead of relying on annotations, PURE elegantly distills robust structural knowledge from state-of-the-art 2D priors into 4D feed-forward architectures.
PURE leverages relative depth constraints, affine-invariant geometry distillation, and pose refinement based on feature correspondences to stabilize geometric predictions under challenging dynamic conditions.
Extensive experiments on both synthetic and real-world dynamic benchmarks demonstrate that PURE delivers consistent improvements across multiple state-of-the-art architectures.
Furthermore, our data-efficient paradigm effectively mitigates domain shifts, successfully unlocking the architectural limits of these base models in challenging, in-the-wild dynamic scenes.
\end{abstract}



\begin{document}

\maketitle

\section{Introduction}

Recovering accurate 3D geometry and camera motion from videos is a fundamental problem in computer vision, with applications~\cite{huang2026pointworld,zhao2024m,gu2025diffusion,zhao2026comovi,burgert2025gowiththeflow} ranging from robotics and autonomous driving to augmented reality and digital content creation. Classical Structure-from-Motion (SfM) and visual SLAM systems achieve high accuracy through iterative optimization over sparse correspondences and bundle adjustment, but rely on rigid-scene assumptions, sufficient parallax, and carefully engineered pipelines. These limitations compromise the robustness of such methods in dynamic scenes and impede their scalability to unconstrained real-world videos.

Recent advances in feed-forward 3D reconstruction models have significantly shifted this paradigm. By leveraging large-scale datasets such as CO3Dv2~\cite{reizenstein2021commonobjects3dlargescale} and powerful Transformer-based~\cite{vaswani2023attentionneed} neural architectures, these models directly regress depth and camera parameters from images or videos, enabling fast and scalable reconstruction. Large video-based models such as VGGT~\cite{wang2025vggt}, Pi3~\cite{wang2025pi3}, and DepthAnything3~\cite{depthanything3} demonstrate impressive zero-shot performance on long sequences, highlighting the potential of data-driven priors for 3D understanding. Despite this progress, most feed-forward models are trained with strong supervision, including ground-truth depth, camera poses, or synthetic renderings, which are costly to acquire and limit generalization to real-world dynamic scenes.

Adapting pre-trained 3D reconstruction models to in-the-wild videos without annotations remains a challenging problem. Existing unsupervised approaches (e.g.,~\cite{godard2019monodepth2, zhou2017sfmlearner}) typically rely on photometric consistency, which fundamentally degrades in the presence of illumination changes, non-Lambertian surfaces, and highly dynamic objects. Conversely, methods relying on test-time optimization or per-scene adaptation~\cite{yuan2025test3rlearningreconstruct3d, li2025megasam, huang2025vipe, lu2025trackingworld} incur prohibitive computational overhead, rendering them impractical for efficient video processing. Consequently, an effective annotation-free strategy that refines feed-forward models at the dataset level---preserving inference efficiency---remains largely underexplored.

In this work, we propose a novel post-training framework, termed \textbf{PURE} (\textbf{P}ost-training \textbf{U}nlabeled \textbf{RE}construction), for annotation-free refinement of feed-forward 3D reconstruction models. PURE adapts a pre-trained base model using unconstrained video sequences, without requiring ground-truth supervision or slow test-time optimization. Instead of relying on photometric signals, PURE distills robust structural knowledge from state-of-the-art 2D monocular priors into the 4D feed-forward architecture. To enable stable refinement despite the absolute scale and shift errors inherent in teacher models, we introduce relative depth constraints that preserve ordinal depth, local affine-invariant geometry distillation that maintains fine-grained surface structure, and correspondence-based pose refinement that enforces multi-view consistency.

\begin{figure*}[t]
    \centering
    \includegraphics[width=\textwidth]{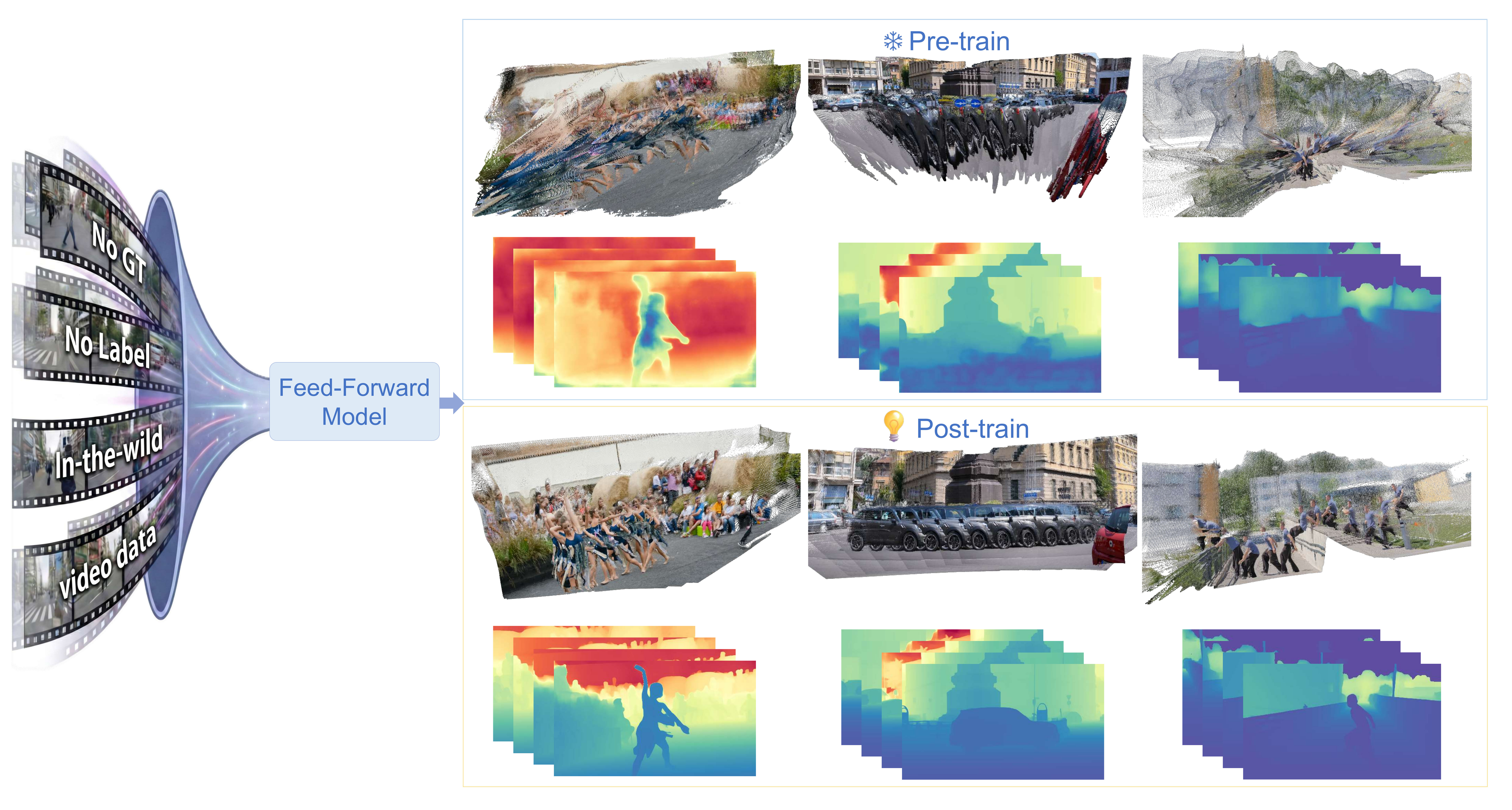}
    \Description{Side-by-side reconstructions before and after unlabeled post-training.}
    \caption{Our annotation-free post-training significantly improves feed-forward 3D reconstruction in dynamic scenes, yielding more accurate and temporally consistent geometry compared to the original pre-trained models.}
    \label{fig:teaser}
\end{figure*}

We validate our approach through extensive experiments on both synthetic and real-world dynamic datasets, including video depth estimation, monocular depth prediction, and camera pose evaluation. Across multiple state-of-the-art architectures, our post-training strategy consistently improves performance and robustness, particularly in challenging dynamic scenes. These results demonstrate that annotation-free post-training offers a practical and scalable pathway for advancing feed-forward 3D reconstruction in real-world settings.

In summary, the contributions of this work are as follows:

\begin{itemize}
\item \textbf{A Novel Post-Training Paradigm:} We propose PURE, a unified, model-agnostic post-training framework that adapts pre-trained feed-forward 3D reconstruction models to in-the-wild videos using purely unlabeled data, eliminating the need for ground-truth annotations or test-time optimization.
\item \textbf{Robust Geometry Distillation:} We introduce a noise-tolerant distillation strategy utilizing relative depth ordering and local affine-invariant constraints. This allows the base model to absorb rich structural priors from 2D teacher models while actively filtering out their absolute scale and shift errors.
\item \textbf{Dynamic-Aware Pose Refinement:} We design a correspondence-based camera pose refinement mechanism featuring a rigorous filtering and sampling pipeline, ensuring reliable multi-view consistency even in scenes dominated by non-rigid dynamic motion.
\end{itemize}
\section{Related Work}

\subsection{Optimization-based 3D Reconstruction}

\begin{figure*}[!ht]
    \centering
    \includegraphics[width=\textwidth]{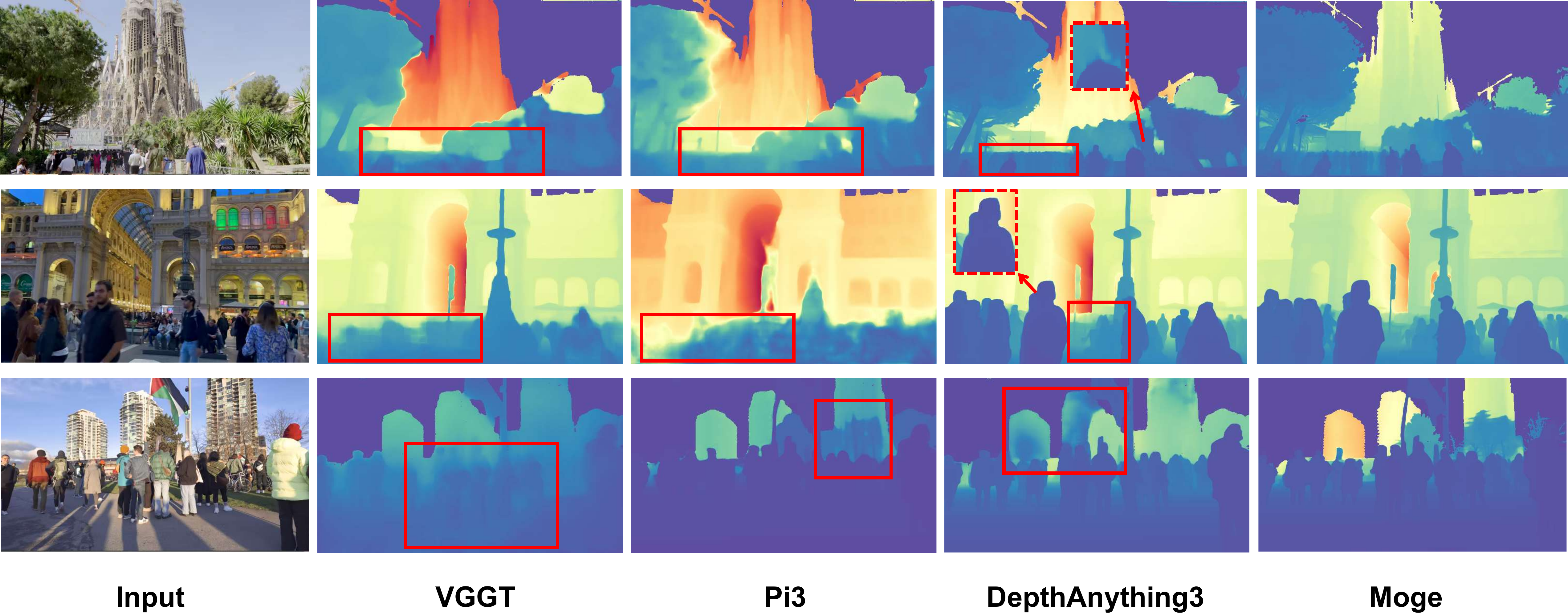}
    \Description{Frame-wise geometry from a monocular teacher compared with a feed-forward model.}
    
    \caption{MoGe provides accurate frame-wise relative geometry, while feed-forward 3D models ensure temporal scale consistency.
We unify both by transferring monocular geometric cues while preserving temporal consistency.
}
    \label{fig:monodepth}
    
\end{figure*}

Classical SfM and visual SLAM systems recover camera poses and scene geometry through iterative optimization using 2D correspondences and bundle adjustment. 
While accurate in controlled settings (e.g., COLMAP~\cite{schoenberger2016sfm, schoenberger2016mvs}), they rely on rigid scenes, sufficient parallax, and reliable feature matches. 

Recent optimization-based methods~\cite{chen2025leapvolongtermeffectivepoint, zhao2022particlesfmexploitingdensepoint, teed2021droid, li2025megasam, huang2025vipe, lu2025trackingworld, wu2025swiftd, wu2025localdygs} extend this paradigm to handle dense geometry and dynamic content. 
Systems such as MegaSAM~\cite{li2025megasam} and ViPE~\cite{huang2025vipe} combine dense depth estimation, motion reasoning, and iterative refinement to reconstruct long video sequences. 
Despite improved robustness, these approaches remain computationally expensive and require careful system design, limiting their scalability to diverse in-the-wild videos.

\subsection{Feed-forward 3D Reconstruction}

To overcome the limitations of iterative optimization, recent works~\cite{wang2024dust3rgeometric3dvision, zhang2024monst3r, lu2024align3r, jang2025pow3rempoweringunconstrained3d, li2025mono3rexploitingmonocularcues, cao2025reconstructing, xiong2025huprior3r, li2026unish, wang2025continuous, streamVGGT, point3r, chen2025ttt3r3dreconstructiontesttime, wang2025vggt, lu2026track4world, wang2025pi3, depthanything3} have explored feed-forward 3D reconstruction models leveraging large-scale data-driven priors. 
These methods can be broadly divided into pair-wise and video-based approaches.

\textbf{Pair-wise methods} originate from DUSt3R~\cite{wang2024dust3rgeometric3dvision}, which formulates multi-view reconstruction as dense point-map regression from image pairs. 
Subsequent works~\cite{zhang2024monst3r, lu2024align3r, jang2025pow3rempoweringunconstrained3d, li2025mono3rexploitingmonocularcues} extend this paradigm to dynamic videos by incorporating temporal fine-tuning or monocular depth priors, but remain limited by pair-wise processing and poor scalability to long sequences.

\textbf{Video-based methods} extend feed-forward reconstruction to full video sequences. 
Models such as CUT3R~\cite{wang2025continuous}, StreamVGGT~\cite{streamVGGT}, Point3R~\cite{point3r}, and TTT3R~\cite{chen2025ttt3r3dreconstructiontesttime} capture long-range temporal dependencies via memory or streaming mechanisms, while VGGT~\cite{wang2025vggt}, Pi3~\cite{wang2025pi3}, and DepthAnything3~\cite{depthanything3} learn strong spatio-temporal priors from large-scale video data. 
Despite strong performance, these models remain brittle on dynamic, in-the-wild videos without annotations, motivating annotation-free post-training refinement such as ours.

\section{Methodology}

\begin{figure*}[t]
    \centering
    \includegraphics[width=\textwidth]{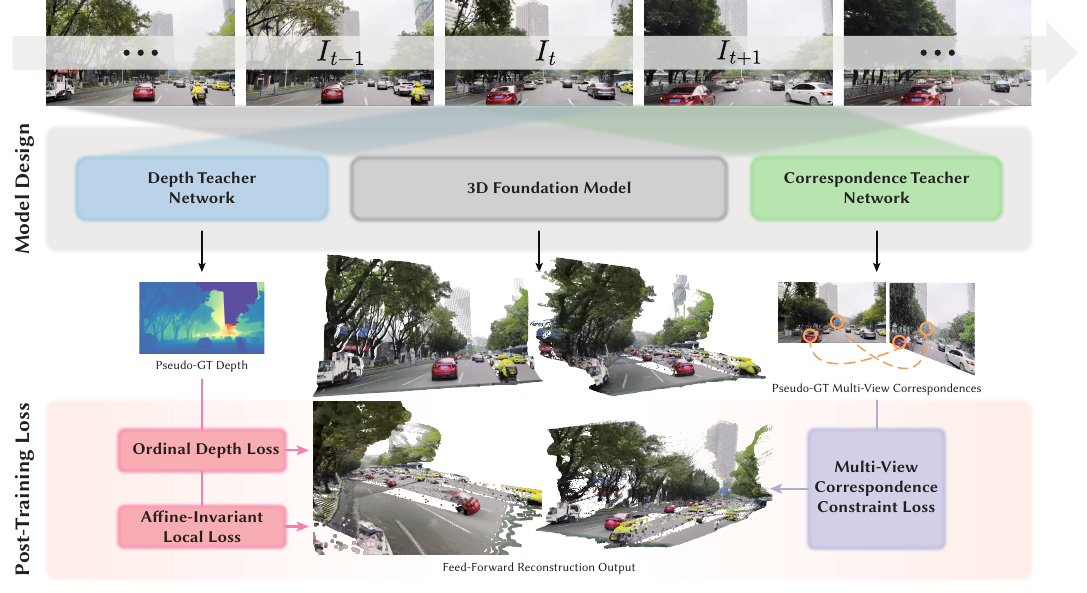}
    \Description{Diagram of the geometry-distillation and correspondence branches in PURE.}
    \caption{
    \textbf{Overview of PURE}, our annotation-free post-training framework.
Given an unlabeled image sequence, a pre-trained feed-forward model predicts depth and camera poses. We refine the predictions via two complementary branches: (1) local geometry refinement using relative depth and affine-invariant constraints with intrinsic distillation; and (2) pose refinement using robust correspondence-based constraints from GIM matches, enforcing cross-frame geometric consistency.
    }
    \label{fig:pipeline}
    
\end{figure*}

We propose a unified framework termed \textbf{PURE} for unlabeled post-training of feed-forward 3D reconstruction models (Fig.~\ref{fig:pipeline}). 
Given a sequence of $K$ images $I = \{I_1, \dots, I_K\}$, the base model $\mathcal{F}_\theta$ predicts camera poses $\mathbf{T}$, intrinsics $\mathbf{K}$, and depth or point maps ($\mathbf{D}$/$\mathbf{P}$). 
These predictions are refined via local geometry optimization using ordinal depth, affine-invariant constraints, and intrinsic distillation, and through dense correspondences from GIM~\cite{shen2024gimlearninggeneralizableimage} filtered by confidence, cycle consistency, and inverse density sampling. 
By combining these mechanisms, our framework improves both geometry and camera pose estimation, yielding more accurate and temporally consistent 3D reconstructions from unlabeled video sequences.

\subsection{Relative Local Geometry Optimization}

Monocular depth estimators~\cite{moge} produce depth maps with more accurate relative ordering and local geometric structures compared to feed-forward 3D reconstruction models~\cite{wang2025vggt,wang2025pi3,depthanything3}, as illustrated in Fig.~\ref{fig:monodepth}. However, their predictions often exhibit temporal scale inconsistencies. Directly supervising feed-forward 3D reconstruction models with monocular depth can lead to pronounced ``layering'' artifacts, since monocular object-wise depth scales may vary significantly across frames and deviate from the 3D reconstruction model’s learned geometric prior.
To address this issue, we avoid absolute depth supervision and instead introduce two relative constraints: an ordinal depth ranking loss and a local affine-invariant loss. These losses transfer the reliable structural and local geometric cues from the monocular teacher while remaining invariant to global scale and shift, enabling stable refinement without collapsing the underlying geometry of the 3D reconstruction models.

\paragraph{\textbf{Ordinal Depth Constraint.}}
For an input image $I_t$, let $D_t$ be the predicted depth map and $\hat{D}_t = \mathcal{T}_{mono}(I_t)$ be the pseudo-ground truth generated by a monocular teacher. Inspired by MoDGS~\cite{liu2025modgsdynamicgaussiansplatting}, we enforce consistency in the relative order of depth values between pixel pairs, making the optimization robust to absolute scale fluctuations. We define the robust ordinal loss as:

\begin{equation}
\begin{split}
\mathcal{L}_{order} = \sum_{(u_1, u_2) \in \mathcal{P}} \bigg\| & \tanh \left( \alpha (D_t(u_1) - D_t(u_2)) \right) \\
& - R(\hat{D}_t(u_1), \hat{D}_t(u_2)) \bigg\|_1
\end{split}
\label{eq:order}
\end{equation}

where $\mathcal{P}$ represents a set of sampled pixel pairs, $\alpha$ is a scaling factor, and $R(\cdot, \cdot)$ is an indicator function returning $\{+1, -1\}$ based on the depth ranking of the teacher model. This constraint preserves structural integrity without propagating erroneous absolute scales.

\paragraph{\textbf{Intrinsic Parameter Distillation.}}

Estimating camera intrinsics---particularly the focal length---from unconstrained videos is inherently ill-posed due to the strong ambiguity between depth scale and intrinsics. 
In practice, we observe that directly aligning 3D point clouds predicted by monocular depth estimators (e.g., MoGe~\cite{moge}) and feed-forward 3D reconstruction models is problematic: discrepancies in the estimated intrinsics result in mismatched ray directions, i.e., the same image pixels are back-projected along different viewing rays, introducing structural distortions that make the resulting point clouds incompatible under simple scale-and-shift 3D alignment.

Interestingly, we find that pre-trained 3D reconstruction models such as VGGT~\cite{wang2025vggt} and Pi3~\cite{wang2025pi3} tend to produce relatively stable intrinsic parameters or ray directions, even when their depth estimates are inaccurate. 
Motivated by this observation, we design a distillation strategy that explicitly decouples ray geometry from depth magnitude.

Building on this insight, we employ a distillation-based strategy to generate pseudo 3D local points. Let $\mathbf{p}$ denote the homogeneous pixel coordinates. We combine the robust ray directions $\mathbf{r}_{pretrain}$, derived from the pre-trained model's intrinsics $\mathbf{K}_{pre}$, with the corresponding depth values $d_{mono}$ from the monocular depth estimator. This process is formulated as:
\begin{align}
\mathbf{r}_{pretrain} &= \mathbf{K}_{pre}^{-1} \mathbf{p} \label{eq:ray} \\
\mathbf{P}_{pseudo} &= d_{mono} \cdot \mathbf{r}_{pretrain} \label{eq:pseudo}
\end{align}
These synthesized 3D points, $\mathbf{P}_{pseudo}$, serve as supervisory signals that implicitly regularize the model's estimation of camera intrinsics, without requiring explicit labels.

\paragraph{\textbf{Local Geometry Constraint.}}
To capture high-frequency surface details, we adopt the Affine-Invariant Distillation Loss from MoGe~\cite{moge}. Instead of aligning the entire image globally---which is sensitive to outliers---we enforce geometric consistency within local spherical regions. For a local region centered at pixel $p_j$, we align the predicted point map $\mathbf{P}$ to the teacher's pseudo point map $\mathbf{P}_{pseudo}$, computed as in \eqref{eq:pseudo}, by solving for the optimal scale $\mathbf{s}$ and shift $\mathbf{t}$:
\begin{equation}
\mathcal{L}_{local} = \sum_{j} \min_{\mathbf{s}, \mathbf{t}} \sum_{i \in \mathcal{N}(p_j)} \| {\mathbf{P}_{pseudo}}_i - (\mathbf{s} \cdot \mathbf{P}_i + \mathbf{t}) \|_1
\label{eq:local}
\end{equation}
This formulation preserves local curvature and fine-grained surface details even under global scale drift.

\subsection{Pose Estimation via Correspondences}

Accurate camera pose estimation is essential for coherent multi-view reconstruction. Traditional Structure-from-Motion (SfM) pipelines rely on sparse 2D feature correspondences to recover epipolar geometry. To approximate this process in an unsupervised manner, we leverage GIM~\cite{shen2024gimlearninggeneralizableimage} for robust correspondence estimation between image pairs. These correspondences serve as powerful constraints, ensuring that matched 2D pixels correspond to the same 3D point in space. To guarantee the reliability of these constraints, we implement a rigorous filtering pipeline:

\begin{itemize}
    \item \textbf{Dynamic and Confidence Filtering:} Matches located in dynamic regions or with confidence scores below a pre-defined threshold $\tau$ are discarded.
    \item \textbf{Cycle Consistency Check:} We enforce forward-backward cycle consistency ($p_A \to p_B \to p_A'$) to remove ambiguous or erroneous matches.
    \item \textbf{Inverse Density Sampling (IDS):} High-confidence correspondences often cluster in texture-rich regions, potentially biasing the global pose estimation. To address this, we employ IDS, which prioritizes sampling matches from sparser, lower-confidence regions. This ensures a spatially uniform distribution of constraints and improves global robustness.
\end{itemize}

\subsection{Model Training}

We fine-tune three representative feed-forward 3D reconstruction models: VGGT~\cite{wang2025vggt}, Pi3~\cite{wang2025pi3}, and DepthAnything3~\cite{depthanything3}. 
For VGGT and DepthAnything3, only a subset of intermediate layers is updated to balance adaptation capability with training stability.
For Pi3, we observe severe convergence instability and grid-like artifacts during fine-tuning. 
We attribute this to its pixel-shuffle--based point head, whose inherent grid structure is amplified by our local geometry constraint. 
To mitigate this, we replace the original point head with a multi-level convolutional feature head, producing smoother and more stable point predictions. 
The new head is jointly trained during post-training.

A key challenge in dynamic scene reconstruction is the lack of high-quality real-world training data. 
Most existing dynamic datasets are synthetic, while real-world datasets such as Waymo~\cite{sun2020scalabilityperceptionautonomousdriving} provide sparse LiDAR depth, which is insufficient for learning fine-grained geometry. 
To address this limitation, we leverage large-scale unlabeled videos from real dynamic scenes. 
As summarized in Tab.~\ref{table:Datasets}, we utilize high-quality sequences from SpatialVID-HQ~\cite{wang2025spatialvidlargescalevideodataset} and Sekai-Real-Walking-HQ~\cite{li2025sekai}. 
Pseudo supervision is obtained using off-the-shelf foundation models: MoGe~\cite{moge} for monocular depth, Grounded SAM 2~\cite{ravi2024sam2segmentimages, liu2023grounding, ren2024grounding, ren2024grounded, kirillov2023segany, jiang2024trex2} for dynamic object and sky segmentation, and GIM Roma~\cite{shen2024gimlearninggeneralizableimage} for dense correspondence estimation.

To further prevent catastrophic forgetting, we augment training with synthetic datasets including OmniWorld~\cite{zhou2025omniworld}, PointOdyssey~\cite{PointOdyssey}, VKITTI~\cite{cabon2020virtualkitti2}, and BEDLAM~\cite{BEDLAM}, as well as real-world driving videos from Waymo. 
This combination of real and synthetic data enables robust post-training across diverse dynamic environments.

\begin{table}[htbp]
    %
    \setlength\tabcolsep{5pt}
    \caption{\textbf{Statistics of datasets for fine-tuning.}}
    
    \label{table:Datasets}
    \centering
    \begin{tabular}{lcccc}
        \toprule
        \multirow{2}{*}{Name} & \multirow{2}{*}{\#Scene}& Avg. & Total  &\multirow{2}{*}{Type} \\
        & & \#Img & \#Img  & \\
        \midrule
        SpatialVID &1.8k&$\sim$490&882k&Real(Unlabeled)\\
        Sekai-Real &4k&900&3.6M&Real(Unlabeled)\\      
        OmniWorld &121&$\sim$4k&484k&Synthetic\\
        VKITTI&100&425&42k&Synthetic\\
        BEDLAM&30&250&7.5k&Synthetic\\
        PointOdyssey &131&1.8k&240k&Synthetic\\
        Waymo&1k&200&200k&Real\\
        \bottomrule
  \end{tabular}
  
\end{table}


\section{Experiments}
\label{sec:experiments}

We present implementation details in Sec.~\ref{sec:implementation} and primarily evaluate our approach on dynamic-scene benchmarks, focusing on video depth estimation (Sec.~\ref{sec:eval-videodepth}) and camera pose estimation (Sec.~\ref{sec:eval-relpose}). To further validate the effectiveness of our annotation-free post-training strategy, we conduct additional experiments in which post-training is performed directly on the test sets (Sec.~\ref{sec:eval-posttraining}). This setting isolates the impact of the proposed post-training objectives from data scale and distribution effects, demonstrating that our method consistently improves geometric predictions even when adaptation is carried out on the evaluation data itself. Finally, we present detailed ablation studies in Sec.~\ref{sec:ablation-study} to analyze the contributions of individual framework components, alongside a scaling-up analysis that investigates the impact of post-training data volume on overall performance.

\begin{table*}[!ht]
    \centering
    \caption{
        \textbf{Video depth estimation on Sintel, Bonn, KITTI and Dycheck.} \textbf{Best} results are highlighted.
    }

    \tablestyle{4.2pt}{1} 
    \begin{adjustbox}{width=\textwidth}
    \begin{tabular}{lcccccccccc}
        \toprule
        \multirow{3}{*}{Method} &
        \multirow{3}{*}{Params} &
        \multirow{3}{*}{Align} &
        \multicolumn{2}{c}{Sintel} &
        \multicolumn{2}{c}{Bonn} &
        \multicolumn{2}{c}{KITTI} &
        \multicolumn{2}{c}{DyCheck} \\ 

        \cmidrule(lr){4-5} \cmidrule(lr){6-7} \cmidrule(lr){8-9} \cmidrule(lr){10-11}
        
        & & &
        Abs Rel $\downarrow$ & $\delta<1.25$ $\uparrow$ &
        Abs Rel $\downarrow$ & $\delta<1.25$ $\uparrow$ &
        Abs Rel $\downarrow$ & $\delta<1.25$ $\uparrow$ &
        Abs Rel $\downarrow$ & $\delta<1.25$ $\uparrow$ \\
        \midrule

        VGGT & 1.16B & \multirow{6}{*}{\makecell[c]{scale}} 
                                          & 0.279 & 0.657 & 0.058 & 0.969 & 0.062 & 0.971 & 0.134 & 0.827 \\
        \textbf{Ours (VGGT)} & 1.16B & & \textbf{0.220} & \textbf{0.724} & \textbf{0.045} & \textbf{0.975} & \textbf{0.040} & \textbf{0.984} & \textbf{0.062} & \textbf{0.956} \\

        \cmidrule(lr){1-1}
        
        Pi3 & 959M & & 0.230 & 0.674 & 0.049 & 0.975 & \textbf{0.038} & 0.986 & \textbf{0.048} & 0.980 \\
        
        \textbf{Ours (Pi3)} & 989M & & \textbf{0.212} & \textbf{0.733} & \textbf{0.043} & \textbf{0.977} & 0.039 & \textbf{0.987} & 0.049 & \textbf{0.981} \\

        \cmidrule(lr){1-1}
        
        DA3& 1.21B & & 0.236 & 0.702 & 0.047 & 0.974 & 0.058 & 0.975 & 0.053 & 0.959\\
        \textbf{Ours (DA3)} & 1.21B & & \textbf{0.224} & \textbf{0.710} & \textbf{0.045} & \textbf{0.975} & \textbf{0.048} & \textbf{0.978} & \textbf{0.040} & \textbf{0.982} \\

        \midrule

        VGGT & 1.16B & \multirow{6}{*}{\makecell[c]{scale \\\&\\ shift}} & 0.215 & 0.704 & 0.050 & 0.971 & 0.053 & 0.969 & 0.113 & 0.852 \\ 
        \textbf{Ours (VGGT)} & 1.16B & & \textbf{0.178} & \textbf{0.741} & \textbf{0.043} & \textbf{0.976} & \textbf{0.038} & \textbf{0.984} & \textbf{0.055} & \textbf{0.976} \\

        \cmidrule(lr){1-1}
        
        Pi3 & 959M & & 0.210 & 0.726 & 0.044 & 0.975 & \textbf{0.037} & \textbf{0.986} & 0.042 & 0.979 \\
        \textbf{Ours (Pi3)} & 989M & & \textbf{0.195} & \textbf{0.750} & \textbf{0.038}  & \textbf{0.977} & 0.038 & 0.987 & \textbf{0.040} & \textbf{0.980} \\

        \cmidrule(lr){1-1}
        
        DA3& 1.21B & & 0.173 & 0.730 & 0.046 & 0.974 & 0.053 & 0.974 & 0.044 & 0.974 \\
        \textbf{Ours (DA3)} & 1.21B & & \textbf{0.156} & \textbf{0.743} & \textbf{0.043} & \textbf{0.976} & \textbf{0.045} & \textbf{0.978} & \textbf{0.036} & \textbf{0.983} \\

        \bottomrule
    \end{tabular}
    
    \label{tab:videodepth}
    \end{adjustbox}
\end{table*}

\begin{table*}[ht]
    \centering
    \caption{
        \textbf{Camera pose estimation on Sintel, Bonn and Dycheck.} 
        We report ATE, and RPE for translation (\textbf{Trans}) and rotation (\textbf{Rot}). \textbf{Best} results are highlighted.
    }
    \tablestyle{4pt}{1.05}
    
    \resizebox{1.5\columnwidth}{!}{
        \begin{tabular}{lccccccccc}
            \toprule
            \multirow{2.5}{*}{Method} &
            \multicolumn{3}{c}{Sintel} &
            \multicolumn{3}{c}{Bonn} &
            \multicolumn{3}{c}{Dycheck} \\
            
            \cmidrule(lr){2-4} \cmidrule(lr){5-7} \cmidrule(lr){8-10}
            
            &
            ATE$\downarrow$ & Trans$\downarrow$ & Rot$\downarrow$ &
            ATE$\downarrow$ & Trans$\downarrow$ & Rot$\downarrow$ &
            ATE$\downarrow$ & Trans$\downarrow$ & Rot$\downarrow$ \\
            \midrule
            
VGGT & 0.148 & 0.044 & 0.455 & 0.015 & 0.010 & 0.633 & 0.022 & 0.007 & 0.672 \\
            \textbf{Ours (VGGT)} & \textbf{0.061} & \textbf{0.039} & \textbf{0.300} & \textbf{0.010} & \textbf{0.010} & \textbf{0.630} & \textbf{0.015} & \textbf{0.005} & \textbf{0.614} \\

            \cmidrule(lr){1-1}
            
            Pi3 & 0.060 & 0.069 & \textbf{0.654} & 0.013 & 0.010 & 0.632 & 0.011 & 0.004 & \textbf{0.613} \\
            \textbf{Ours (Pi3)} & \textbf{0.058} & \textbf{0.066} & 0.668 & \textbf{0.013} & \textbf{0.010} & \textbf{0.630} & \textbf{0.010} & \textbf{0.004} & 0.615 \\

            \cmidrule(lr){1-1}
            
            DA3 & 0.078 & 0.046 & 0.336 & \textbf{0.008} & \textbf{0.008} & \textbf{0.626} & 0.007 & 0.003 & 0.619 \\        
            \textbf{Ours (DA3)} & \textbf{0.052} & \textbf{0.030} & \textbf{0.205} & 0.009 & 0.009 & 0.628 & \textbf{0.006} & \textbf{0.003} & \textbf{0.610} \\
            \bottomrule
        \end{tabular}
    }
    \label{tab:pose_dist}
\end{table*}

\begin{figure*}[!ht]
    \centering
    
    \includegraphics[width=\textwidth]{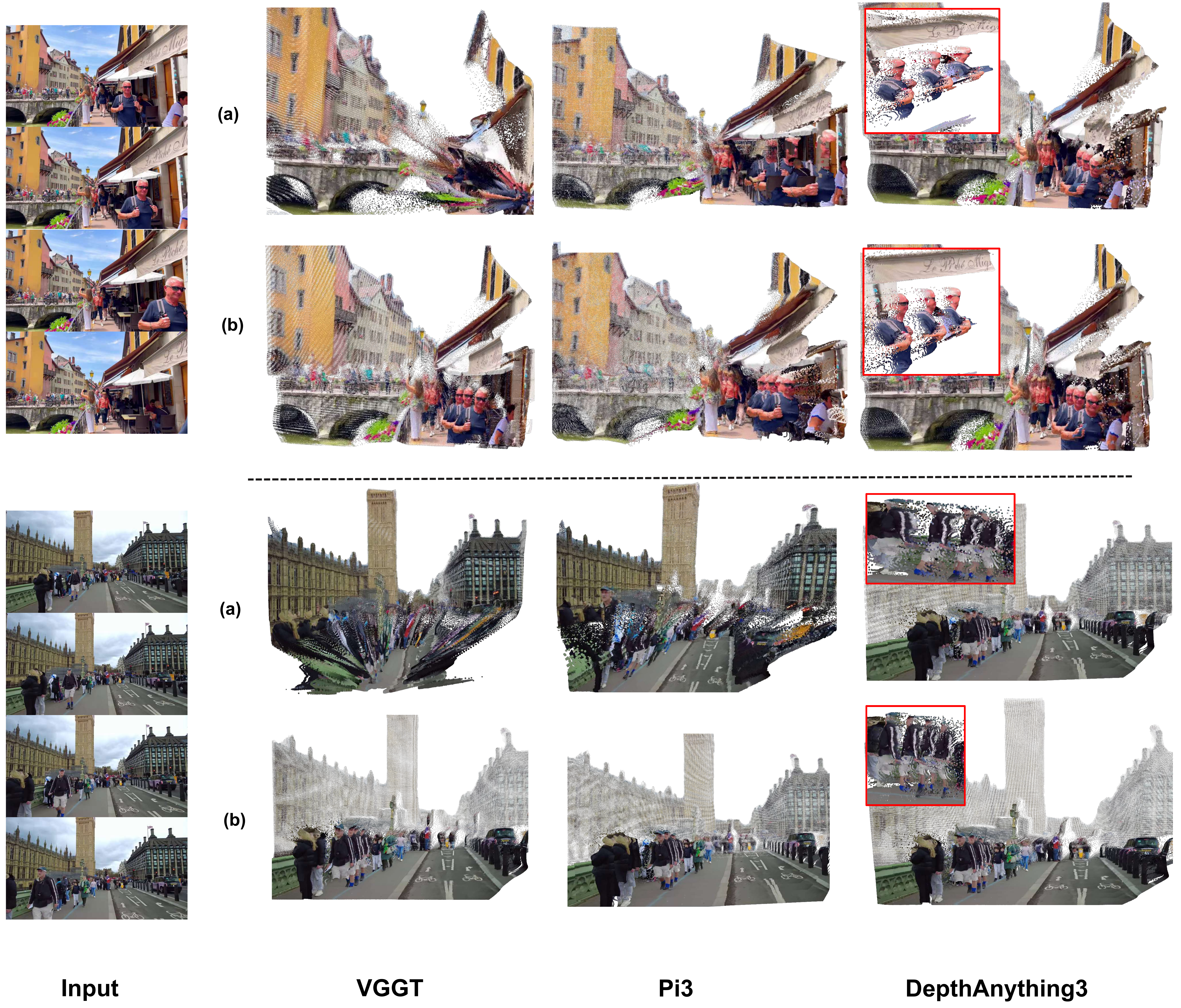}
    \Description{Qualitative reconstructions before and after post-training on dynamic scenes.}
    
    \caption{~\textbf{Qualitative comparison results.} The first row (a) shows the reconstruction results produced by the original models, while the second row (b) presents the reconstructions after post-training with our proposed method, exhibiting improved geometric consistency and reconstruction quality, particularly in dynamic regions.}
    \label{fig:dynamic}
    
\end{figure*}

\subsection{Implementation Details}\label{sec:implementation}

We fine-tune three baseline models: VGGT~\cite{wang2025vggt}, Pi3~\cite{wang2025pi3}, and 
DepthAnything3 (DA3)~\cite{depthanything3}. For VGGT and DA3, optimization is applied to the intermediate global attention layers. For Pi3, in addition to these layers, we also train a newly replaced point prediction head.

To support video inputs with varying resolutions, we constrain image aspect ratios to the range $[0.5, 2.0]$ and limit the total number of pixels to between $160\text{k}$ and $250\text{k}$. Our training setup includes both multi-view and single-view inputs, where the number of views is uniformly sampled from $[1, 24]$. To ensure stable training efficiency, the batch size is dynamically adjusted to maintain an approximately constant number of tokens per iteration.

All experiments are conducted on 4$\times$ H800 GPUs, using dynamic batch sizing and a gradient accumulation step of 2. We adopt AdamW optimizer with a learning rate of $1\times10^{-5}$, weight decay of $5\times10^{-2}$, and $\beta=(0.9, 0.95)$. The learning rate is scheduled using OneCycleLR with cosine annealing and a maximum learning rate of $1\times10^{-5}$. Gradient clipping is applied with a maximum norm of 1.0.

In terms of training efficiency, the PURE post-training stage takes approximately 20 hours for VGGT (12K iterations) and 8 hours for DA3 (5K iterations). For Pi3, training is divided into two stages: a depth head initialization phase (10K iterations, approximately 12 hours), followed by a post-training phase (5K iterations, approximately 8 hours).

Notably, PURE is a post-training framework that introduces negligible architectural modifications (except for a minor head replacement in Pi3). As a result, the inference computational cost remains identical to that of the original base models.

\subsection{Video Depth Estimation}\label{sec:eval-videodepth}

We evaluate our method on video depth estimation in dynamic scenes, which is particularly challenging due to non-rigid motion, moving objects, and temporal inconsistencies. Following prior work~\cite{wang2025continuous, wang2025vggt, wang2025pi3}, we conduct experiments on a diverse set of real-world and synthetic datasets, including Bonn~\cite{bonn}, Sintel~\cite{sintel}, KITTI~\cite{kitti}, and Dycheck~\cite{dycheck}, and additionally present qualitative results on in-the-wild videos.

We report Absolute Relative Error (Abs Rel) and threshold accuracy ($\delta<1.25$) under two alignment settings: (i) scale-only alignment and (ii) joint scale and 3D translation alignment. 
As shown in Tab.~\ref{tab:videodepth}, 
our proposed post-training consistently improves performance across nearly all datasets and backbone architectures, demonstrating the robustness of our approach.
Notably, even for Pi3---where the original pixel-shuffle point head is replaced and retrained with fewer parameters---our method yields clear gains. 
VGGT and DepthAnything3 also benefit consistently from post-training across all evaluated datasets, indicating effective geometric refinement.

As illustrated in Fig.~\ref{fig:dynamic}, all three models exhibit visibly improved reconstruction quality after post-training, especially in regions with strong object motion. 
Post-training enhances temporal consistency and reduces dynamic artifacts, leading to more stable and accurate depth predictions. 
These qualitative results corroborate the quantitative improvements and highlight the effectiveness of our annotation-free post-training strategy for dynamic scene reconstruction.

\subsection{Camera Pose Estimation}\label{sec:eval-relpose}

We evaluate predicted camera poses using two complementary sets of metrics: angular accuracy~\cite{wang2025vggt, wang2025pi3} and distance error~\cite{zhang2024monst3r, wang2025continuous}. 
Specifically, we measure performance using the Absolute Trajectory Error (ATE), Relative Pose Error for translation (RPE$_\text{trans}$), and Relative Pose Error for rotation (RPE$_\text{rot}$) on both synthetic and real-world datasets. 

Experiments are conducted on the synthetic outdoor Sintel~\cite{sintel} dataset, as well as the real-world indoor Bonn~\cite{bonn} and DyCheck~\cite{dycheck} datasets. 
To account for scale ambiguity, all predicted camera trajectories are first aligned to the corresponding ground-truth trajectories using a Sim(3) transformation before computing errors.
As shown in Tab.~\ref{tab:pose_dist}, post-training with our annotation-free strategy consistently improves camera pose estimation accuracy across all datasets, demonstrating that our method enhances robustness to non-rigid motion and complex scene dynamics.

\subsection{Post-Training with Test-Set Data}\label{sec:eval-posttraining}

To evaluate the effectiveness of our annotation-free post-training strategy under limited data regimes, we conduct additional experiments on VGGT~\cite{wang2025vggt} using images from the Bonn~\cite{bonn}, Sintel~\cite{sintel}, and KITTI~\cite{kitti} test sets. 
These images are processed and used for post-training with our proposed framework.

For the Sintel dataset, we post-train on a selected subset, excluding scenes with insufficient static regions for reliable correspondence estimation. Following standard protocols, we report Abs Rel for depth under joint scale and 3D translation alignment, and ATE for camera pose estimation.

As shown in Tab.~\ref{tab:post_training_ablation}, post-training on test-set data consistently improves performance across all evaluated benchmarks. 
However, the gains are smaller than those obtained when post-training is conducted on large-scale unlabeled video datasets. 
This indicates that, while the proposed method remains effective even with limited data, its benefits scale more substantially with increased data diversity and volume. 
These results underscore the importance of scaling up annotation-free post-training on rich, large-scale unlabeled video collections to fully unlock the potential of feed-forward 3D reconstruction models.

{\setlength{\belowcaptionskip}{-8pt}
\begin{table}[t]
\centering
\small
\caption{\textbf{Effect of annotation-free post-training on VGGT.}
We report depth and camera pose performance before and after post-training on three benchmarks.}
\label{tab:post_training_ablation}
\resizebox{\columnwidth}{!}{
\begin{tabular}{lcccc}
\toprule
\multirow{2}{*}{Dataset} &
\multicolumn{2}{c}{Depth (Abs Rel $\downarrow$)} &
\multicolumn{2}{c}{Camera Pose (ATE $\downarrow$)} \\
\cmidrule(lr){2-3} \cmidrule(lr){4-5}
& Before & After & Before & After \\
\midrule
Sintel  & 0.157 & \textbf{0.147} & 0.048 & \textbf{0.037} \\
Bonn    & 0.052 & \textbf{0.041} & 0.015 & \textbf{0.012} \\
KITTI   & 0.054 & \textbf{0.049} & - & - \\
\bottomrule
\end{tabular}
}

\end{table}
}

\subsection{Ablation Study}\label{sec:ablation-study}

To thoroughly analyze the contribution of each component in our annotation-free post-training framework, we conduct ablation studies on the overall framework effectiveness and its scaling behavior with increasing data.

\paragraph{\textbf{Component Ablation.}}
We first evaluate the impact of each component on the DyCheck dataset using VGGT as the base model. All variants are tested under the same protocol, reporting both depth estimation accuracy and camera trajectory errors.

Specifically, \emph{w/o all} denotes a variant trained only on datasets with ground-truth annotations (Tab.~\ref{table:Datasets}), without leveraging our large-scale unlabeled video data. This results in clear performance degradation, demonstrating the critical importance of annotation-free data for both depth and pose estimation.

We further ablate individual loss terms. Removing the ordinal depth loss ($\mathcal{L}_{order}$) degrades both depth accuracy and camera stability, indicating its importance in preserving relative depth ordering in unconstrained scenes. Similarly, removing the local geometry distillation loss ($\mathcal{L}_{local}$) leads to worse performance, particularly in camera trajectory metrics, highlighting its role in enforcing cross-view geometric consistency.

Overall, the full model achieves the best performance across all metrics, confirming that both large-scale unlabeled data and the proposed geometric supervision losses are essential and complementary.

{\setlength{\belowcaptionskip}{-8pt}
\begin{table}[t]
    \centering
    \small
    \caption{
        \textbf{Ablation study on Dycheck dataset.} 
        We evaluate the impact of different loss terms on depth estimation and camera trajectory accuracy.
    }

    \tabcolsep=4pt
    \resizebox{1.0\columnwidth}{!}{

        \begin{tabular}{lcc c ccc}
            \toprule
            \multirow{2}{*}{Method} & \multicolumn{2}{c}{Depth } & & \multicolumn{3}{c}{Camera Pose} \\
            \cmidrule(lr){2-3} \cmidrule(lr){5-7}
            & Abs Rel $\downarrow$ & $\delta < 1.25$ $\uparrow$ & & ATE $\downarrow$ & Trans $\downarrow$ & Rot $\downarrow$ \\
            \midrule
            Base Model           & 0.113 & 0.852  & & 0.022 & 0.007 & 0.672  \\
            w/o all & 0.084 & 0.892 & & 0.019 & 0.006 & 0.656 \\
            w/o $\mathcal{L}_{order}$  & 0.062 & 0.957 &  & 0.018 & 0.005 & 0.635 \\
            w/o $\mathcal{L}_{local}$  & 0.075 & 0.935 & & 0.016 & 0.005 & 0.640 \\ 
            \midrule
            \textbf{Ours (Full)} & \textbf{0.055} & \textbf{0.976} & & \textbf{0.015} & \textbf{0.005} & \textbf{0.614} \\
            \bottomrule
        \end{tabular}
    }
    \label{tab:ablation_dycheck}
\end{table}
}

\begin{figure}[htbp]
    \centering
    \includegraphics[width=\columnwidth]{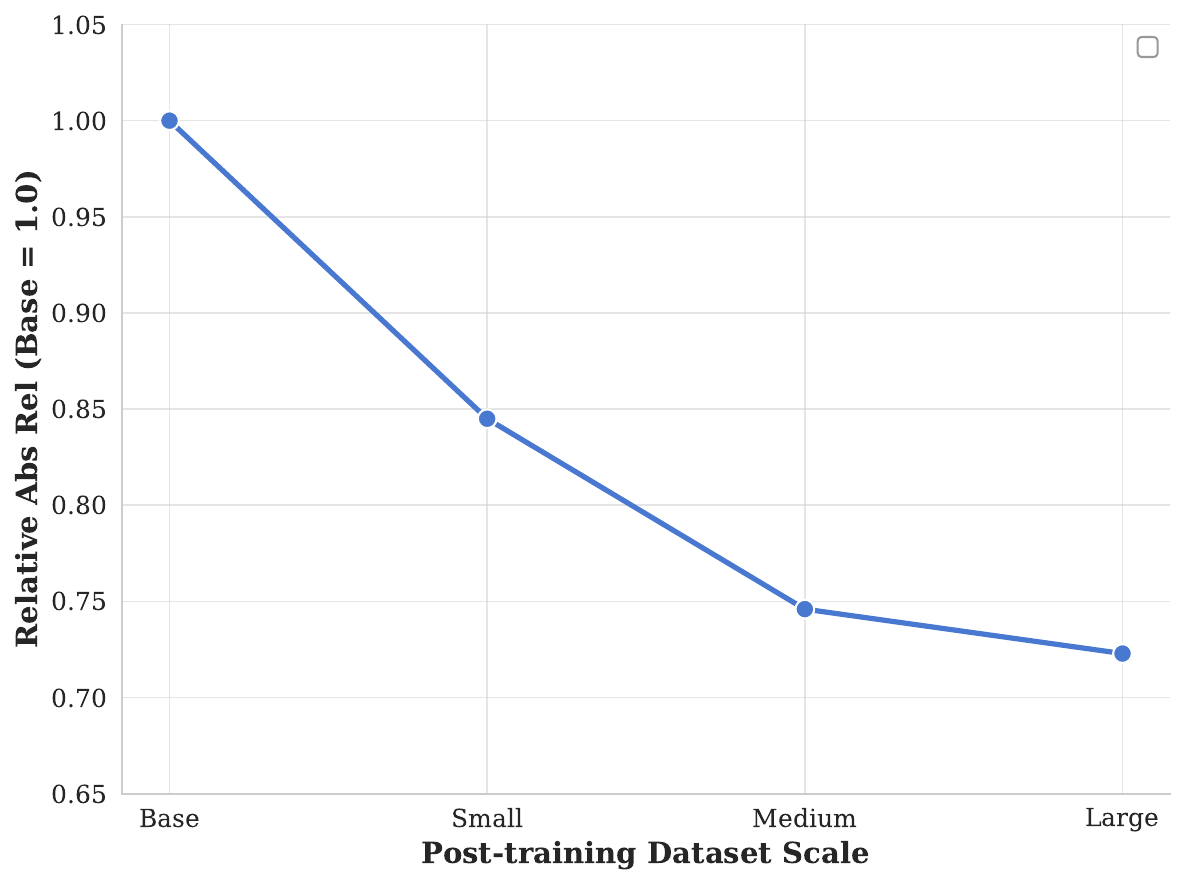}
    \Description{Bar chart of relative depth error as the unlabeled training set grows.} 
    \caption{
        \textbf{Scaling behavior with increasing post-training data.}
        Performance improves significantly as more unlabeled data is introduced, but gradually saturates at larger scales due to the fixed model capacity.
    }
    \label{fig:scaling_ablation}
\end{figure}

\paragraph{\textbf{Scaling-Up Analysis.}}
We further study how performance scales with increasing amounts of post-training data. We construct three dataset scales: Small ($\sim$0.2M images), Medium ($\sim$1M images), and Large ($\sim$5M images). 

We report the average Absolute Relative Error (Abs Rel) across four datasets (Sintel, Bonn, KITTI, and DyCheck), normalized by the base model (set to 1.0). As shown in Fig.~\ref{fig:scaling_ablation}, performance improves significantly at smaller scales (0.8452 for Small and 0.7461 for Medium), but saturates at larger scales (0.7229 for Large).

This saturation is expected, as PURE is a post-training framework that does not increase model capacity. Instead, it unlocks the latent potential of the base model by exposing it to diverse, in-the-wild data. The results demonstrate that while scaling data is effective, gains are ultimately bounded by the underlying model architecture.

\section{Conclusion}

We present \textbf{PURE}, an annotation-free post-training framework for adapting feed-forward 3D reconstruction models to unconstrained dynamic videos. By distilling relative geometric priors and enforcing correspondence-based pose consistency, PURE significantly enhances depth and camera estimation across state-of-the-art architectures. Ultimately, this work demonstrates that unsupervised post-training is a highly effective, scalable pathway for unlocking robust 3D reconstruction in the wild.

\bibliographystyle{ACM-Reference-Format}
\bibliography{references}

\clearpage
\appendix

\section{Training Details}

\subsection{Details of Local Geometry Loss}

To ensure robustness against outliers in the local geometry, our alignment loss $\mathcal{L}_{local}$ employs the $L_1$ norm (Manhattan distance) rather than the standard $L_2$ norm. The objective is to find the optimal scale $\mathbf{s}^*$ and shift $\mathbf{t}^*$ that minimize the discrepancy between the predicted local structure $\mathbf{P}$ and the pseudo-ground truth $\mathbf{P}_{pseudo}$:

$$(\mathbf{s}^*, \mathbf{t}^*) = \arg\min_{\mathbf{s}, \mathbf{t}} \sum_{i \in \mathcal{N}(p_j)} \| {\mathbf{P}_{pseudo}}_i - (\mathbf{s} \cdot \mathbf{P}_i + \mathbf{t}) \|_1$$

Unlike least-squares optimization, this $L_1$ minimization problem does not possess a direct closed-form solution for simultaneous scale and shift estimation. To solve this efficiently during the training forward pass, we adopt a candidate hypothesis strategy based on the optimality conditions of the $L_1$ norm.

Since the optimal solution for $L_1$ minimization typically interpolates at least one data point (i.e., the residual is zero for some $k \in \mathcal{N}(p_j)$), we iterate through all points in the neighborhood to generate a set of candidate affine transformations. The procedure is as follows:Hypothesis Generation: For each element $k$ in the local neighborhood $\mathcal{N}(p_j)$, we hypothesize that the optimal shift $\mathbf{t}_k$ aligns the $k$-th element of $\mathbf{P}$ perfectly with $\mathbf{P}_{pseudo}$. This imposes a constraint that reduces the problem dimension.Conditional Scale Optimization: Given the alignment at index $k$, the optimal scale $\mathbf{s}_k$ that minimizes the $L_1$ error over the remaining points can be computed efficiently (equivalent to solving a weighted median problem on the ratios).Selection: We compute the total alignment error for each candidate pair $(\mathbf{s}_k, \mathbf{t}_k)$ and select the pair that yields the global minimum:$$(\mathbf{s}^*, \mathbf{t}^*) = \arg\min_{k} \left( \sum_{i} \| {\mathbf{P}_{pseudo}}_i - (\mathbf{s}_k \cdot \mathbf{P}_i + \mathbf{t}_k) \|_1 \right)$$Finally, we use the optimal $(\mathbf{s}^*, \mathbf{t}^*)$ to compute the loss value. Gradients are backpropagated through the selected parameters to update the network, treating the selection of the optimal candidate as a deterministic step.

\subsection{Details of Inverse Density Sampling (IDS)}

High-confidence correspondences generated by matching networks often exhibit strong spatial clustering, typically concentrating in high-texture areas. This violates the assumption of spatially independent and identically distributed (i.i.d.) noise required by many geometric solvers (e.g., PnP with RANSAC), leading to biased pose estimation. To mitigate this, we propose Inverse Density Sampling (IDS), which reformulates the selection process as a weighted sampling problem.

\paragraph{Density Estimation}

We model the spatial distribution of the candidate correspondence set $\mathcal{M} = \{(\mathbf{u}_i, \mathbf{u}'_i, s_i)\}_{i=1}^N$ using Kernel Density Estimation (KDE). Here, $\mathbf{u}_i$ denotes the 2D coordinates in the query image, and $s_i$ is the predicted confidence score. The local spatial density $\rho_i$ for the $i$-th match is defined as:

$$\rho_i = \frac{1}{Z} \sum_{j \neq i} K\left( \frac{\| \mathbf{u}_i - \mathbf{u}_j \|_2}{\sigma} \right)$$

where $K(\cdot)$ is a Gaussian kernel $K(x) = \exp(-x^2)$, $\sigma$ is the bandwidth parameter controlling the neighborhood size (set proportional to the image resolution), and $Z$ is a normalization constant. A high $\rho_i$ indicates that the $i$-th match is located in a dense cluster of candidates.

\paragraph{Sampling Probability}
The goal of IDS is to select a subset of matches that maximizes spatial coverage while maintaining high reliability. We define the sampling weight $w_i$ for each correspondence as

$$w_i = \frac{f(s_i)}{\rho_i + \epsilon}$$

where:

\begin{itemize}

\item The denominator $\rho_i + \epsilon$ penalizes matches in over-represented regions ($\epsilon$ is a small constant for numerical stability).

\item The numerator $f(s_i)$ ensures that we do not sample outliers solely because they are isolated. We typically use $f(s_i) = s_i^\alpha$ (with $\alpha \ge 1$), which acts as a high-pass filter for confidence.

\end{itemize}

The final selection probability $P(i)$ is obtained by normalizing the weights: $P(i) = w_i / \sum_k w_k$. By sampling according to $P(i)$, the network effectively "flattens" the spatial histogram of the correspondences, ensuring that constraints from sparse but informative regions are utilized in the pose solver.

\section{More Results}

\subsection{Additional Quantitative Results}
As shown in Tables~\ref{tab:videodepth_sub} and ~\ref{tab:pose_sub}, PURE consistently improves the performance of all three adapted base models, VGGT, PI3, and DA3. The gains are particularly pronounced for VGGT, which we attribute to its relatively limited training data. By leveraging large-scale unlabled in-the-wild video data, PURE effectively expands the training distribution and better unlocks the capacity of the VGGT architecture, leading to substantial performance improvements.

Notably, PURE not only enhances these existing models but also surpasses other state-of-the-art 4D reconstruction methods across both depth estimation and camera pose benchmarks, demonstrating its effectiveness as a scalable and general post-training framework.

\begin{table*}[!ht]
    \centering
    \caption{
        Video depth estimation on Sintel, Bonn, KITTI and Dycheck. Best results are \textbf{bolded} and second best are \underline{underlined}. Average ranks across all 8 metrics are reported in the last column.
    }
    
    \tablestyle{4.2pt}{1} 
    \begin{adjustbox}{width=\textwidth}
    \begin{tabular}{lccccccccc}
        \toprule
        \multirow{3}{*}{Method} &
        \multicolumn{2}{c}{Sintel} &
        \multicolumn{2}{c}{Bonn} &
        \multicolumn{2}{c}{KITTI} &
        \multicolumn{2}{c}{DyCheck} & 
        \multirow{3}{*}{\textbf{Avg Rank} $\downarrow$} \\ 

        \cmidrule(lr){2-3} \cmidrule(lr){4-5} \cmidrule(lr){6-7} \cmidrule(lr){8-9}
        
        & Abs Rel $\downarrow$ & $\delta<1.25$ $\uparrow$ &
        Abs Rel $\downarrow$ & $\delta<1.25$ $\uparrow$ &
        Abs Rel $\downarrow$ & $\delta<1.25$ $\uparrow$ &
        Abs Rel $\downarrow$ & $\delta<1.25$ $\uparrow$ & \\
        \midrule

        GeometryCrafter\cite{xu2025geometrycrafterconsistentgeometryestimation} & 0.311 & 0.685 & 0.085 & 0.932 & 0.099 & 0.939 & 0.098 & 0.918 & 7.9 \\
        Geo4D\cite{jiang2025geo4dleveragingvideogenerators} & 0.205 & 0.735 & 0.059 & 0.972 & 0.086 & 0.937 & 0.081 & 0.934 & 6.1 \\
        Vipe\cite{huang2025vipe} & 0.560 & 0.452 & 0.108 & 0.865 & 0.178 & 0.765 & 0.128 & 0.796 & 10.0 \\ 
        TTT3R\cite{chen2025ttt3r3dreconstructiontesttime} & 0.469 & 0.510 & 0.061 & 0.969 & 0.151 & 0.818 & 0.117 & 0.823 & 8.7 \\ 

        VGGT\cite{wang2025vggt} & 0.215 & 0.704 & 0.050 & 0.971 & 0.053 & 0.969 & 0.113 & 0.852 & 6.8 \\ 
        Pi3\cite{wang2025pi3} & 0.210 & 0.726 & 0.044 & 0.975 & \textbf{0.037} & \underline{0.986} & 0.042 & 0.979 & 3.6 \\
        DA3\cite{depthanything3} & \underline{0.173} & 0.730 & 0.046 & 0.974 & 0.053 & 0.974 & 0.044 & 0.974 & 4.5 \\

        \cmidrule(lr){1-1}
        \textbf{Ours (VGGT)} & 0.178 & 0.741 & \underline{0.043} & \underline{0.976} & \underline{0.038} & 0.984 & 0.055 & 0.976 & 3.2 \\
        \textbf{Ours (Pi3)} & 0.195 & \textbf{0.750} & \textbf{0.038} & \textbf{0.977} & \underline{0.038} & \textbf{0.987} & \underline{0.040} & \underline{0.980} & \textbf{1.8} \\        
        \textbf{Ours (DA3)} & \textbf{0.156} & \underline{0.743} & \underline{0.043} & \underline{0.976} & 0.045 & 0.978 & \textbf{0.036} & \textbf{0.983} & \underline{2.2} \\

        \bottomrule
    \end{tabular}
    
    \label{tab:videodepth_sub}
    \end{adjustbox}
\end{table*}

\begin{table*}[!ht]
    \centering
    \caption{
        \textbf{Camera pose estimation on Sintel, Bonn and Dycheck.} 
        We report ATE, and RPE for translation (\textbf{Trans}) and rotation (\textbf{Rot}). The last column shows the \textbf{Average Rank} across all metrics and datasets ($\downarrow$ indicates lower is better). Best results are \textbf{bolded} and second best are \underline{underlined}.
    }
    
    \tablestyle{3.5pt}{1} 
    \begin{adjustbox}{width=0.9\textwidth}
    \begin{tabular}{lcccccccccc}
        \toprule
        \multirow{3}{*}{Method} &
        \multicolumn{3}{c}{Sintel} &
        \multicolumn{3}{c}{Bonn} &
        \multicolumn{3}{c}{DyCheck} &
        \multirow{3}{*}{\textbf{Avg Rank} $\downarrow$} \\ 

        \cmidrule(lr){2-4} \cmidrule(lr){5-7} \cmidrule(lr){8-10}
        
        & ATE $\downarrow$ & Trans $\downarrow$ & Rot $\downarrow$ &
        ATE $\downarrow$ & Trans $\downarrow$ & Rot $\downarrow$ &
        ATE $\downarrow$ & Trans $\downarrow$ & Rot $\downarrow$ & \\
        \midrule

        Geo4D\cite{jiang2025geo4dleveragingvideogenerators} & 0.192 & 0.055 & 0.510 & 0.022 & 0.015 & 0.683 & 0.031 & 0.012 & 0.705 & 6.6 \\
        Vipe\cite{huang2025vipe} & 0.241 & 0.088 & 0.757 & 0.045 & 0.028 & 0.850 & 0.060 & 0.025 & 0.916 & 8.6 \\ 
        TTT3R\cite{chen2025ttt3r3dreconstructiontesttime} & 0.280 & 0.090 & 0.850 & 0.035 & 0.022 & 0.750 & 0.045 & 0.018 & 0.810 & 8.3 \\ 

        VGGT\cite{wang2025vggt} & 0.148 & 0.044 & 0.455 & 0.015 & 0.010 & 0.633 & 0.022 & 0.007 & 0.672 & 5.3 \\ 
        Pi3\cite{wang2025pi3} & 0.060 & 0.069 & 0.654 & 0.013 & 0.010 & 0.632 & 0.011 & 0.004 & \underline{0.613} & 4.4 \\
        DA3\cite{depthanything3} & 0.078 & 0.046 & 0.336 & \textbf{0.008} & \textbf{0.008} & \textbf{0.626} & \underline{0.007} & \textbf{0.003} & 0.619 & \underline{2.6} \\

        \cmidrule(lr){1-1} 
        \textbf{Ours (VGGT)} & 0.061 & \underline{0.039} & \underline{0.300} & \underline{0.010} & \underline{0.010} & 0.630 & 0.015 & 0.005 & 0.614 & 3.5 \\
        \textbf{Ours (Pi3)} & \underline{0.058} & 0.066 & 0.668 & 0.013 & \underline{0.010} & 0.630 & 0.010 & 0.004 & 0.615 & 4.2 \\        
        \textbf{Ours (DA3)} & \textbf{0.052} & \textbf{0.030} & \textbf{0.205} & 0.009 & 0.009 & \underline{0.628} & \textbf{0.006} & \textbf{0.003} & \textbf{0.610} & \textbf{1.4} \\

        \bottomrule
    \end{tabular}
    
    \label{tab:pose_sub}
    \end{adjustbox}
\end{table*}

\subsection{Additional Qualitative Results}

\begin{figure*}[!ht]
    \centering
    
    \includegraphics[width=\textwidth,height=0.9\textheight,keepaspectratio]{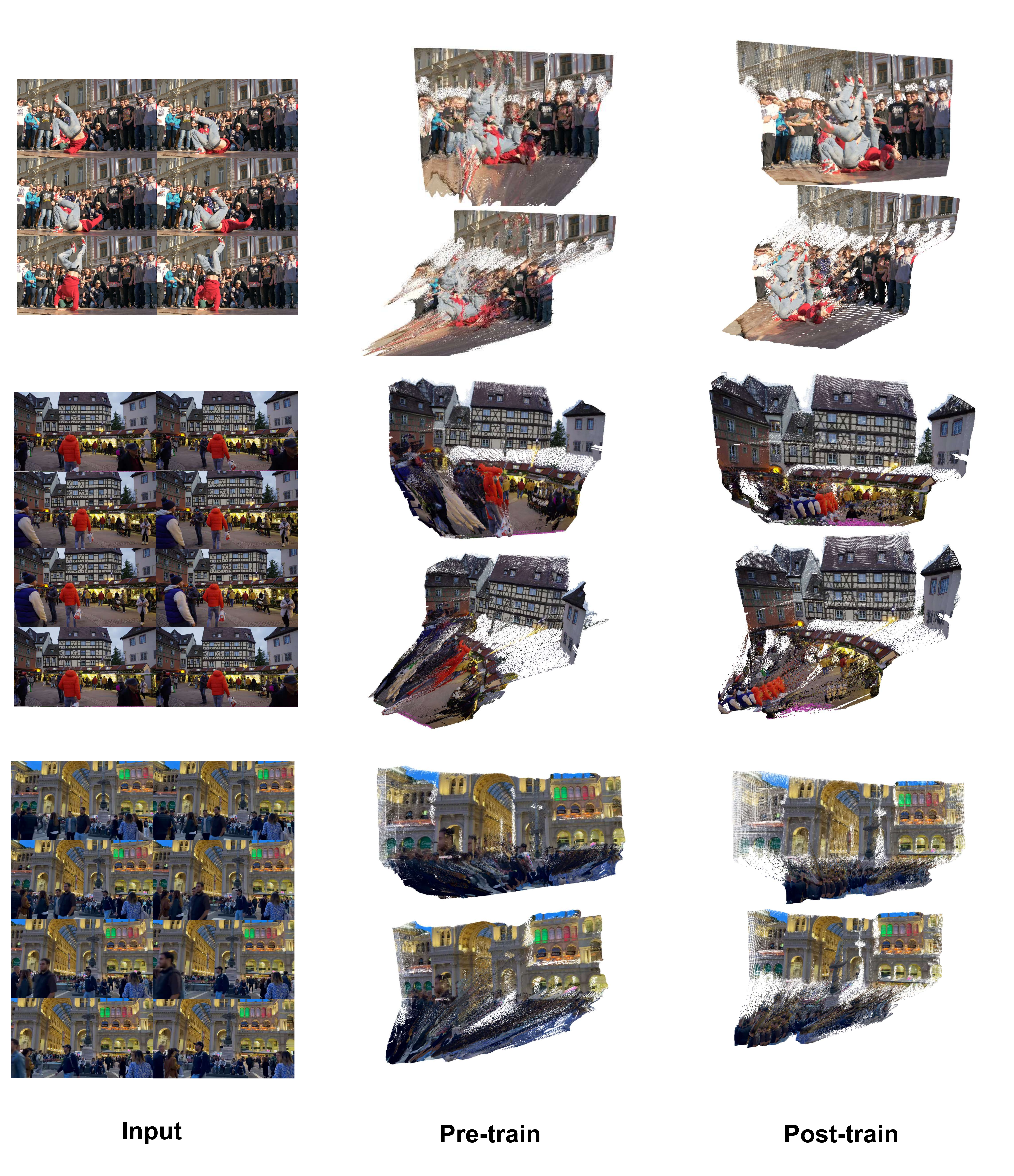}
    
    \caption{Qualitative comparisons before and after applying PURE on diverse in-the-wild videos. PURE significantly improves depth accuracy and sharpens structural details.}
    \Description{Comparisons of geometry before and after PURE on diverse real-world videos, showing improved depth accuracy and sharper structural details.}
    \label{fig:QualitativeResults}
    
\end{figure*}

As further illustrated in Fig.~\ref{fig:QualitativeResults}, PURE yields more accurate geometry with sharper structural details across diverse real-world scenes.

\subsection{Correspondence Estimation Ablation}
We analyze the effectiveness of key components in our correspondence estimation pipeline through qualitative comparisons.

\begin{figure*}[t]
    \centering
    \includegraphics[width=0.8\textwidth]{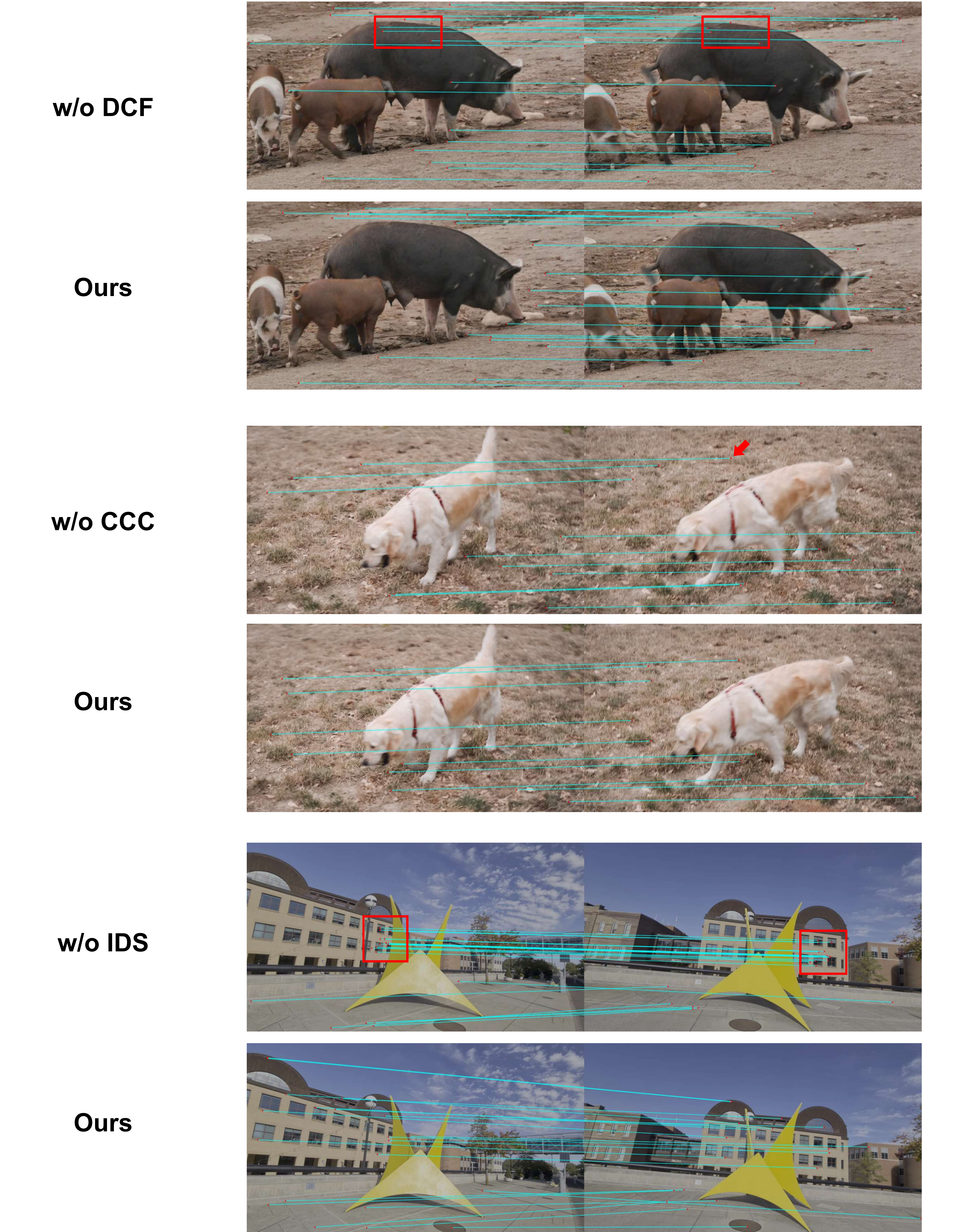}
    \caption{
        \textbf{Ablation of correspondence estimation components.}
        Removing DCF leads to incorrect matches on dynamic regions, CCC improves matching robustness, and IDS ensures uniform spatial coverage.
        Each component is essential for reliable geometric supervision.
    }
    \label{fig:correspondence_ablation}
    \Description{Correspondence comparisons with filtering, cycle consistency, or inverse density sampling removed, illustrating their effects on matching reliability and spatial coverage.}
\end{figure*}

As shown in Fig.~\ref{fig:correspondence_ablation}, removing Dynamic and Confidence Filtering (DCF) causes the model to select points on dynamic objects, leading to corrupted camera pose estimation. Without Cycle Consistency Check (CCC), matching robustness degrades, producing clear mismatches. Without Inverse Density Sampling (IDS), correspondences become overly concentrated in high-confidence regions, resulting in insufficient geometric coverage of the scene.

These results demonstrate that each component plays a crucial role in ensuring accurate, robust, and well-distributed correspondences.

\section{Limitations}

While the PURE framework significantly enhances temporal stability and geometric consistency, it presents a few inherent limitations.

PURE relies on reliable correspondence matching to construct geometric constraints. Performance naturally degrades under conditions that disrupt this matching process, such as abrupt motions causing heavy occlusion and blur, or scenes overwhelmingly dominated by dynamic objects. In such extreme cases, valid static matches plummet, and PURE must fall back entirely on the base model's pre-trained priors, limiting the effectiveness of our geometric refinement.

As a post-training distillation framework, PURE does not expand the base model's parameter capacity. As demonstrated in our scaling experiments, performance gains eventually saturate at the base network's theoretical limit. Furthermore, while our robust constraints effectively filter out absolute scale and shift errors, the overall performance ceiling remains inherently tied to the structural knowledge bounds of the chosen teacher models.

\begin{figure*}[!ht]
    \centering
    
    \includegraphics[width=\textwidth]{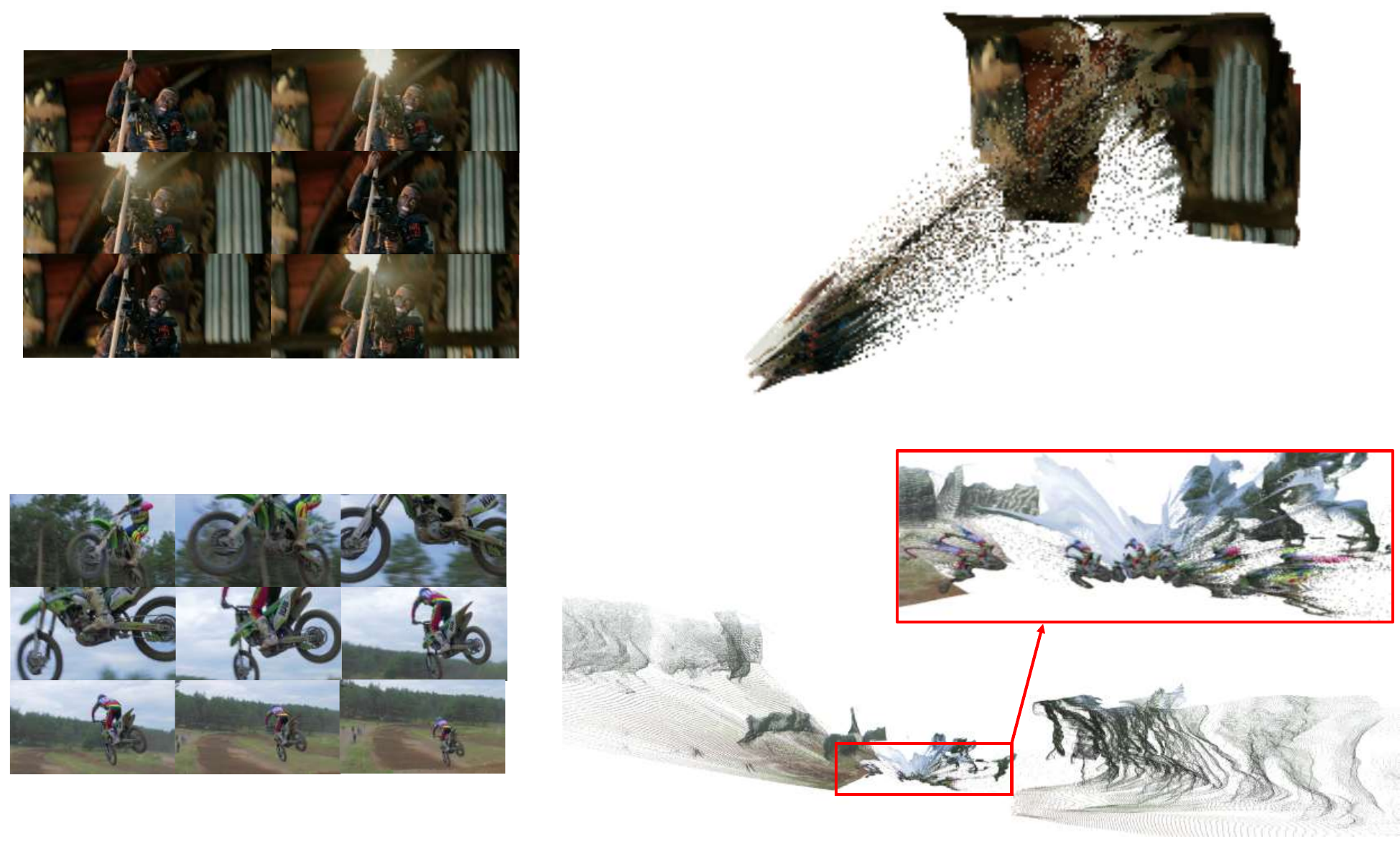}
    
    \caption{Failure cases.}
    \Description{Examples of failure cases for PURE.}
    \label{fig:failure_cases}
    
\end{figure*}

\section{Future Work}

Future work includes exploring alternative geometric supervisions to address severe motion blur and highly dynamic scenes where traditional matching breaks down. We also plan to incorporate higher-capacity base and teacher models into the PURE pipeline to further push performance limits. Given sufficient compute and large-scale data, scaling up via training larger models from scratch represents a promising direction to unlock additional gains.

\end{document}